\documentclass[letterpaper, 10 pt, conference]{ieeeconf}  % Comment this line out if you need a4paper

\IEEEoverridecommandlockouts                              % This command is only needed if 
\usepackage{epsfig}
\usepackage{amsmath}
\usepackage{amssymb}
\usepackage{cite}
\usepackage{subcaption}

\title{\LARGE \bf
Design of the First Insect-scale Spinning-wing Robot
}

\author{Palak Bhushan$^{*}$ and Claire Tomlin$^{*}$% <-this % stops a space
\thanks{$^{*}$The authors are with the Department of EECS, University of California, Berkeley, CA 94720, USA.
        {\tt\footnotesize palak@berkeley.edu} \text{(corresponding author)}, {\tt\footnotesize tomlin@berkeley.edu}.}%
}

\begin{document}

\maketitle
\thispagestyle{empty}
\pagestyle{empty}

%%%%%%%%%%%%%%%%%%%%%%%%%%%%%%%%%%%%%%%%%%%%%%%%%%%%%%%%%%%%%%%%%%%%%%%%%%%%%%%%
\begin{abstract} 
Here we present the design of an insect-scale microrobot that generates lift by spinning its wings. 
This is in contrast to most other microrobot designs at this size scale which rely on flapping wings to produce lift. 
The robot has a wing span of 4 centimeters and weighs 133 milligrams. 
It spins its wings at 47 revolutions/second generating $>$ 138 milligrams of lift while consuming approximately 60 milliwatts of total power and operating at a low voltage ($<$ 3 V). 
Of the total power consumed 8.8 milliwatts is mechanical power generated, part of which goes towards spinning the wings, and 51 milliwatts is wasted in resistive Joule heating. 
With a lift-to-power ratio of 2.3 grams/W, its performance is at par with the best reported flapping wing devices at the insect-scale. 
\end{abstract} 

%%%%%%%%%%%%%%%%%%%%%%%%%%%%%%%%%%%%%%%%%%%%%%%%%%%%%%%%%%%%%%%%%%%%%%%%%%%%%%%%
\section{Introduction} 

Many autonomous flying robots have been built starting from the gram-scale \cite{delfly} and up, 
but no autonomous flyer has yet been built at the 100mg mass-scale. 
Apart from the general difficulty in the construction and assembly at small scales, and unavailability of small off-the-shelf components like batteries,  
a more pronounced difficulty lies in the fundamental physics of scaling. 
For larger sized flying bots electromagnetic (EM) rotary motors are a good option since they can provide high mechanical power output per unit mass and still have very low Joule heat loss in their windings. 
However, EM motors scale down badly \cite{scaling}. With $s$ being the linear scaling rate and mass scaling as $s^3$, if heat loss per unit volume and the motor revolution speed are kept constant, then the motor power output scales as $s^4$ and thus the mechanical power density scales as $s^1$. 

There are many metrics to quantize the performance of actuators like mechanical power density and efficiency, but these can be misleading. For example, an actuator can have a high mechanical power density by simply increasing the motor current but can be very power inefficient and vice-a-versa. Hence for flying robots we use a more informative performance metric which is the ratio of the lift produced (in grams) and the total power consumed (in Watts). 

Given the ineffective scaling of EM motors, other kind of actuators have been used at the milligram-scale to power flight 
with the most popular being piezoelectric actuators \cite{best_PZT}. 
Among all the different locomotion strategies flight, and especially hovering, is the most power demanding \cite{actuator_selection}. 
It is then no surprise that only two flying bots have been reported yet that can lift-off without tethers \cite{robofly18, xwing19}. 
Robofly \cite{robofly18} weighs 190mg and consumes 300mW, most of which is to power its 100mg power electronics circuit that drives its 200V piezoelectric actuator. This gives its lift-to-power ratio as 0.6g/W. 
The newer RoboBee X-wing \cite{xwing19} weighs 260mg and consumes 120mW again mostly to power its 90mg power circuit instead of its 200V piezoelectric actuator, and has a lift-to-power ratio of 2.4g/W. 
Even though they are able to demonstrate untethered operation for a split-second, the problem is that they consume a lot of power and need heavy power electronics making them impractical for future micro-batteries and leaving less mass-budget for useful payloads. 

\begin{figure}
\centering
\epsfig{file=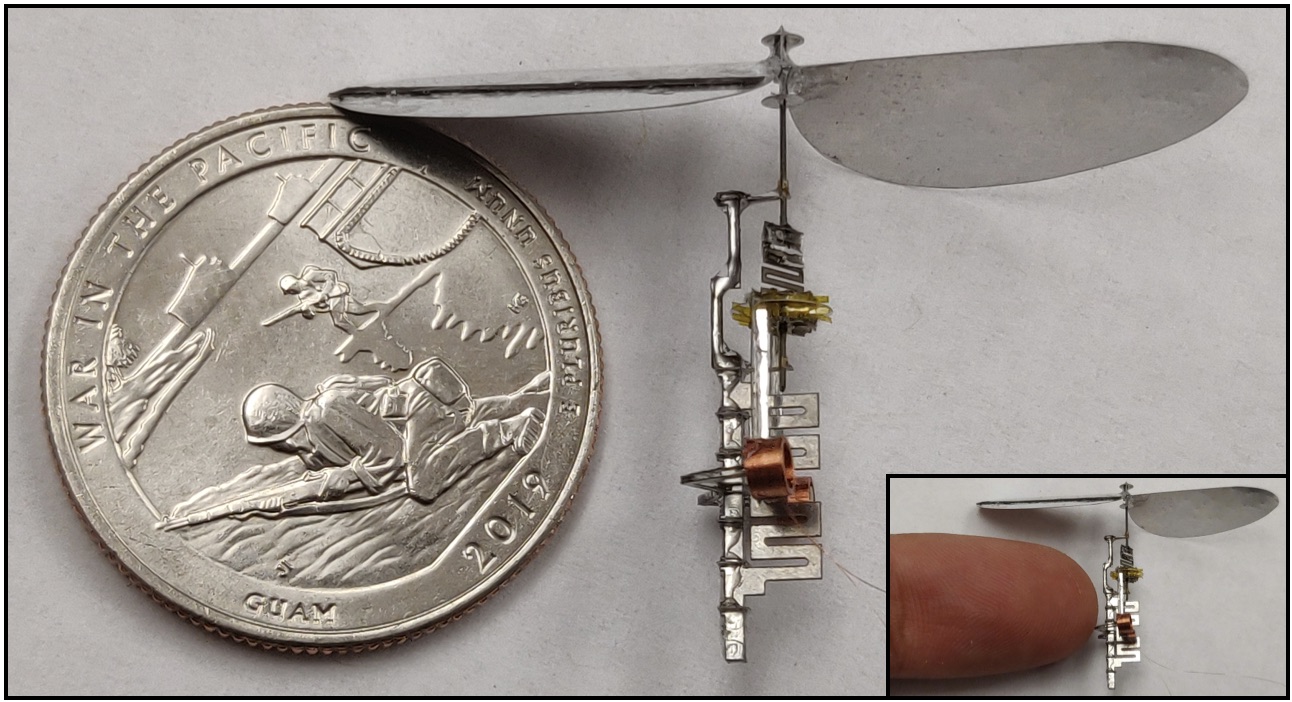,width=3.4in}
\vspace{-1.8em}
\caption{\small{Spinning wing bot compared with a quarter dollar, and, an index finger (in the inset).}}
\label{fig:main-pic}
\end{figure}

Piezoelectric actuators themselves are very efficient in terms of electrical power consumed to mechanical power produced but their performance is brought down due to the high drive voltage demands. 
Here we plan on using low-voltage EM motors to eliminate the use of any inefficient and heavy voltage step-up circuits. 
We choose our coil impedance in order to power it directly using low-voltage power sources. 
However, it can still consume high power by drawing high current like the EM flyers designed in \cite{zhang16,lin_EM1,lin_EM2} -- these actuators are low-voltage but consume more than a watt to operate, giving their lift-to-power ratio $<$ 0.1g/W. 

In order to utilize EM actuators, we need to improve their efficiency. 
One way to accomplish this is frequency-downsampling (or, gearing down), that is, operating the actuator at a high frequency but driving the wings at a low/normal frequency. This reduces the coil forces and thus the coil currents for generating the same amount of mechanical power, thus reducing resistive heat loss. 
Some ideas for gearing down have been used previously for microrobots \cite{wobble1}. 
Here we use a ratchet mechanism \cite{rolling_ral19, jumping_ral19} to convert the motion of high-frequency small-displacement actuators to a low-frequency continuous rotation motion, in part similar to how inchworm motors add up small motions \cite{inchworm02}. 

Another strategy used to improve system efficiency is the use of spinning wings as opposed to flapping wings to generate lift. 
Things in nature don't tend to have a continuous rotation revolute joint possibly due to the problem of supplying blood through veins to the rotating part (though there are exceptions like the flagellum of a bacteria \cite{flagellum}). 
Thus nature came up with flapping wing motion as a means to push air downwards while only performing limited rotations. 
Almost all research works on insect-scale flying robots go the biomimetic way and use flapping wings. 
However, computational studies have shown that spinning wings, if made, are more power efficient than flapping wings \cite{lentink16}. This means that spinning consumes lower power to generate the same amount of lift when compared to flapping motion, and is thus the approach taken here.  

%%%%%%%%%%%%%%%%%%%%%%%%%%%%%%%%%%%%%%%%%%%%%%%%%%%%%%%%%%%%%%%%%%%%%%%%%%%%%%%%
\section{Design} 

\begin{figure}[h]
\centering
\epsfig{file=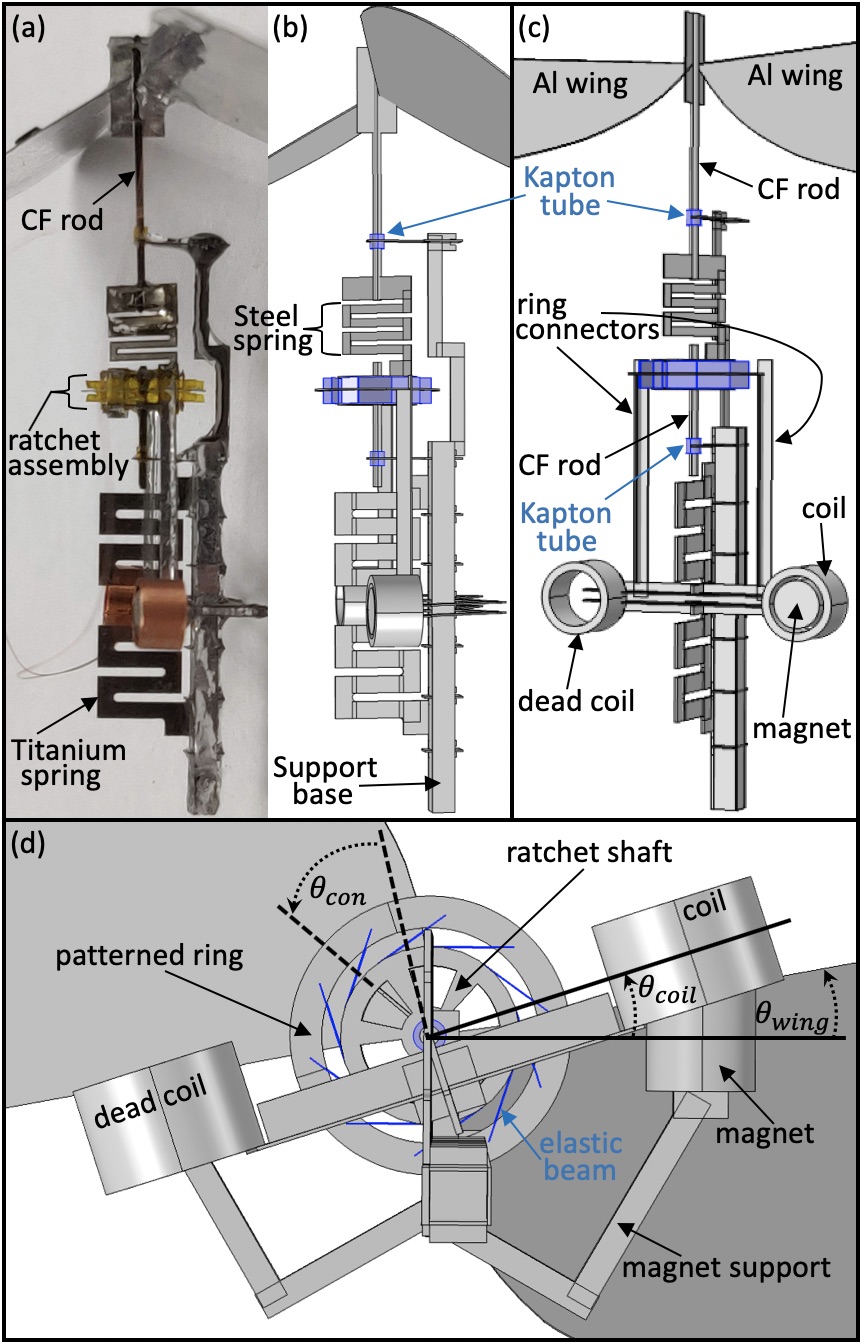,width=3.4in}
\vspace{-1.8em}
\caption{\small{Different components of the spinning wing bot.}}
\label{fig:parts}
\end{figure}

The goal is to convert the oscillatory motion of an actuator to a continuous constant speed rotation in order to spin the wings. We summarize the basic building blocks of our design for accomplishing this as follows. 
\begin{itemize}
\item The actuator is an electromagnetic system comprising of a magnet and a coil (see Fig. \ref{fig:parts}(c), and \S\ref{sec:actuator}). The coil is driven at resonance using a Titanium torsion spring (see Fig. \ref{fig:parts}(a), and \S\ref{sec:Ti-spring}), and mechanical power is supplied to this system by running a current through the coil. 
\item The wings are heavy and form a high rotational inertia system thus acting like a flywheel. This flywheel always rotates anti-clockwise when seen from the bottom (see Fig. \ref{fig:parts}(d), and \S\ref{sec:flywheel}).
\item The oscillatory kinematics of the resonant actuator is used to drive a ratchet which transmits torque, and thus mechanical power, to the flywheel in only the anti-clockwise direction (see Fig. \ref{fig:parts}(d), and \S\ref{sec:ratchet}). The ratchet ensures that the actuator disengages from the flywheel when the actuator moves in the clockwise direction. 
\item The ratchet is connected to the flywheel via a Steel torsion spring (see Fig. \ref{fig:parts}(b), and \S\ref{sec:Steel-spring}). This spring compensates for the speed mismatch between the actuator (which has a periodic angular speed) and the flywheel (which has a constant angular speed). 
\item The flywheel acts as a buffer for kinetic energy and smoothes out the intermittent torque input from the actuator. It ensures that the wings spin at a near-constant angular speed. 
\end{itemize}

%%%%%%%%%%%%
\subsection{Aerodynamic power} 

The total lift and drag forces experienced by the two wings can be described by \cite{passive_rot} 
\begin{equation} \label{eqn:FLFD}
\begin{split}
F_L = ~& 2 \cdot \frac{1}{2} \rho_{air} A (\dot\theta_{wing, ss}\hat p R)^2 C_L, \\
F_D = ~& 2 \cdot \frac{1}{2} \rho_{air} A (\dot\theta_{wing, ss}\hat p R)^2 C_D
\end{split}
\end{equation}
where $\rho_{air}=1.22$kg/m$^3$ is the density of air, $A = R\cdot \frac{R}{A_r}$ is the area of each wing, $R$ is the length of a single wing and $A_r $ is its aspect ratio, $\dot\theta_{wing, ss} = 2\pi f_{wing, ss}$ and $f_{wing, ss}$ is the steady-state revolution rate of the wings, and $\hat p R$ is the radial distance from the spin axis of the center of pressure of a single wing where all the aerodynamic forces are assumed to act upon. $C_L$ \& $C_D$ are the lift and drag coefficients, respectively, of each wing and are given by \cite{passive_rot}
\begin{equation} \label{eqn:CLCD}
\begin{split}
C_L = ~& 1.8 \sin(2 \alpha),\\
C_D = ~& 1.9 - 1.5 \cos(2 \alpha)
\end{split}
\end{equation}
where $\alpha$ is the angle of attack of each wing. The wing assembly is fabricated as per the following parameters 
\begin{equation} \label{eqn:parms}
R = 20\text{mm}, A_r = 4, \alpha = 30^\circ 
\end{equation}
and $\hat p$ is estimated as $=0.46$ during experimentation. The aerodynamic power required to overcome the drag can be given as 
\begin{equation} \label{eqn:Paero}
P_{aero} = F_D \cdot (\dot\theta_{wing, ss}\hat p R).
\end{equation}
For a steady-state speed of $f_{wing, ss} = 47$rev/s the lift, drag and aerodynamic power required are computed to be 
\begin{equation*} \label{eqn:}
F_{L,ss} = 1.4\text{mN, } F_{D,ss} = 1.0\text{mN, and, } P_{aero} = 2.8\text{mW}.
\end{equation*}

%%%%%%%%%%%%
\subsection{Electromagnetic Actuator} \label{sec:actuator}

The details of the actuator design can be found in \cite{baybug18, baybug19}.
Our actuator is based on the Lorentz force and is made up of a magnet and a coil (see Fig. \ref{fig:parts}(d)). Even though the coil moves along a circular arc of radius $r=4$mm, we approximate its path with a linear trajectory parametrized by $y=r\cdot \theta_{coil} = y_{max} \sin(2 \pi f_{coil} t)$ for the analysis in this section.  
The coil is custom made from a 25$\mu$m-thin Copper wire which is array wound $n_{turns} = 48\times8$ number of times. It has an inner diameter of 1.9mm, an outer diameter of 2.45mm, and a height of 1.6mm, with resistance $R_{coil}\approx 108\Omega$. The NdFeB magnet is grade N52 with 1.6mm diameter and height. $y = 0$mm marks the neutral position of the coil when it is perfectly concentric with the magnet. 
The force experienced by the coil can be given by 
\begin{equation} \label{eqn:Fcoil}
F_{coil}(y,t) = B(y)\cdot I_{coil}(y,\dot y, t) \cdot l_{coil} \cdot n_{turns}
\end{equation}
where $B(y)$ is the effective magnetic field seen by the coil at a displacement = $y$ and is computed via FEA, $I_{coil}(y,\dot y, t)$ is the current in the coil, and $l_{coil} = 2\pi r_{coil, avg}$ is the average circumference of the coil with $r_{coil, avg} = 2.2$mm being its average radius. 
The coil is driven externally by a square-wave voltage of amplitude $= \pm V_{max}$ and frequency $= 2f_{coil}$ written as  
\begin{equation} \label{eqn:Vs}
V_s(t) = V_{max} \cdot \text{square}(2f_{coil}\cdot t). 
\end{equation}
This supply voltage frequency is double that of the coil's oscillation frequency since there are 2 pole faces on the magnet and the coil traverses each pole face twice every cycle. 
The sinusoidal motion of the coil through the magnet generates a back-emf in the coil given by 
\begin{equation} \label{eqn:Vemf}
V_{emf}(y,\dot y) = B(y)\cdot l_{coil} \cdot n_{turns}\cdot \dot y
\end{equation}
and thus the coil current can be given by 
\begin{equation} \label{eqn:Icoil}
I_{coil}(y,\dot y, t) = (V_s(t) - V_{emf}(y,\dot y))/R_{coil}. 
\end{equation}
The mechanical power $P_{mech}$ generated by the coil, the resistive heat loss $P_{heat}$, and the total power $P_{net}$ supplied to the coil can now be written as 
\begin{equation} \label{eqn:Pcoil}
\begin{split}
& P_{mech}  = V_{emf} \cdot I_{coil}, \\ 
& P_{heat}  = I_{coil}^2 \cdot R_{coil}, \\ 
& P_{net}  = P_{mech} + P_{heat} = V_s(t) \cdot  I_{coil}. 
\end{split}
\end{equation}

\begin{figure}[h]
\centering
\epsfig{file=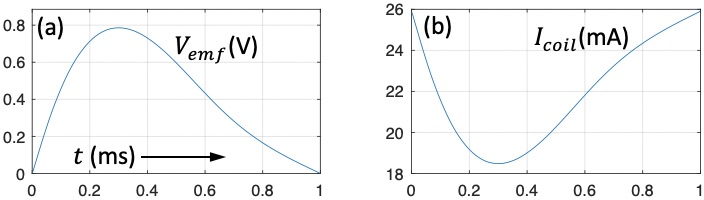,width=3.4in}
\vspace{-1.8em}
\caption{\small{Sample waveforms of the (a) back-emf, and, (b) current produced in the coil in a quarter-cycle.}}
\label{fig:Vemf}
\end{figure}

For $f_{coil} = 250$Hz and $y_{max} = 1.8$mm (corresponding to $\theta_{coil} = \pm 26^\circ$ oscillation), the back-emf generated in the coil is shown in Fig. \ref{fig:Vemf}(a). For $V_{max} = 2.75$V the coil current produced is shown in Fig. \ref{fig:Vemf}(b). 

For the above values of $f_{coil}$, $y_{max}$ and $V_{max}$ the average values of power terms in the coil are computed to be 
\begin{equation*} \label{eqn:}
P_{mech}^{avg} = 8.8\text{mW, } P_{heat}^{avg} = 51\text{mW, and, } P_{net}^{avg} = 59.8\text{mW}.
\end{equation*}
Note that the mechanical power generated in the coil is higher than that consumed by the aerodynamic damping $P_{aero} = 2.8$mW. 
This is because the rest of the mechanical power ($=6$mW) goes towards other mechanical losses, for example, the friction losses in the ratchet and the revolute joints, and the aerodynamic damping experienced by the coil. 

%%%%%%%%%%%%
\subsection{Titanium torsion spring} \label{sec:Ti-spring}

A torsion spring, as shown in Fig. \ref{fig:parts}(a), is used to move the coil along a periodic circular trajectory. 
The spring plus coil system is driven at resonance to produce the coil's kinematics. 
A diametrically opposite dead coil (see Figs. \ref{fig:parts}(c) \& (d)) is used to balance the system to avoid any off-axis forces on the spring and thus eliminate any parasitic motions.  
The desired spring stiffness $k_{coil}$ is determined by the target $f_{coil} = 250$Hz and the system inertia $I_{coil} = 2\cdot m_{coil} \cdot r^2 = 2 \cdot \text{(13mg)} \cdot \text{(4mm)}^2 \Rightarrow k_{coil} = I_{coil} \cdot (2 \pi f_{coil})^2 \approx 1100\mu$Nm. 
The following values are assumed for the spring material which is Ti6Al4V. 
\begin{equation*}
Y = 114\text{GPa}, \epsilon_{max} = 0.43\%, \rho = 4500 \text{kg/m}^3
\end{equation*}
where $Y$ is the Young's modulus, $\epsilon_{max}$ is the fatigue limit, and $\rho$ is the density. 
The procedure for spring design found in \cite{baybug18} is used to dimension the spring, and the result is reported in Table \ref{tab:Ti-spring}. 

\begin{table}[h]
\normalsize
  \centering 
    \caption{\small Specifications of the Titanium torsion spring.}
    \vspace{-0.4em}
    \label{tab:Ti-spring}
    \begin{tabular}{|l|r|}
      \hline
      \multicolumn{2}{|c|}{Spring parameters}\\ 
      \hline
      Max. allowable rotation (before failure) & $36^\circ$ \\
      Torsional stiffness $k_{coil}$ & 1100$\mu$Nm \\
        \hline
      \multicolumn{2}{|c|}{Spring dimensions}\\ 
      \hline 
      Number of parallel beam segments $N$ & 8 \\  
      Beam length $l$ & 1.83mm \\ 
      Beam width $w$ & 0.4mm \\ 
      Beam thickness $t$ & 100$\mu$m\\     
      \hline
    \end{tabular}
\end{table}

%%%%%%%%%%%%
\subsection{Flywheel energy} \label{sec:flywheel}

The wings are laser cut from a 50$\mu$m-thick Aluminum sheet and each wing weighs $\approx 20$mg. 
Approximating the wings by a rod of length $2R$ and mass $40$mg, the rotational inertia $I_{wing}$ of the wing assembly can be approximated as $\frac{1}{12}\cdot (2\cdot20\text{mg}) \cdot (2R)^2 = 5333$mg$\cdot$mm$^2$.
At $f_{wing,ss}=47$rev/s speed this stores kinetic energy $E_{kinetic}= \frac{1}{2}I_{wing}(2 \pi f_{wing,ss})^2 = 233\mu$J. 
If a mechanical power = 8.8mW always leaks out of this system, then there are two cases described as follows depending on the whether any mechanical power is supplied to the flywheel. 
\begin{itemize}
\item In the absence of any input power to the flywheel, mechanical power loss rate = 8.8mW, and thus the kinetic energy drops by 8.8mW$\cdot$2ms $=17.8\mu$J, that is, by 7.6\% in a time interval of 2ms $\Rightarrow $ its speed drops by 3.8\% in 2ms. 
\item If by some means $2\times17.8\mu$J of energy is added to the flywheel, then it gains a net energy of $2\times17.8\mu$J - $17.8\mu$J = $17.8\mu$J, and its speed rises by 3.8\%. 
\end{itemize}

If the flywheel is switched back-and-forth between the above 2 states (that is, if mechanical power is supplied to the flywheel intermittently), 
then its speed will vary by $\pm 1.9$\% around the mean value of $f_{wing,ss}=47$rev/s with a time period of 4ms = a frequency of 250Hz. 

%%%%%%%%%%%%
\subsection{Ratchet} \label{sec:ratchet}

In the absence of any mechanical power input to the flywheel, its speed will continue to decrease until it drops to zero. 
In order to transmit power from the actuator to the flywheel a ratchet is used. 
The ratchet is a mechanism that transmits torque (and thus mechanical power) from the actuator to the flywheel only when the coil moves in the anti-clockwise direction. 
We describe the ratchet here briefly, but we refer the reader to \cite{rolling_ral19} to find the details about its working and construction. 

\begin{figure}[h]
\centering
\epsfig{file=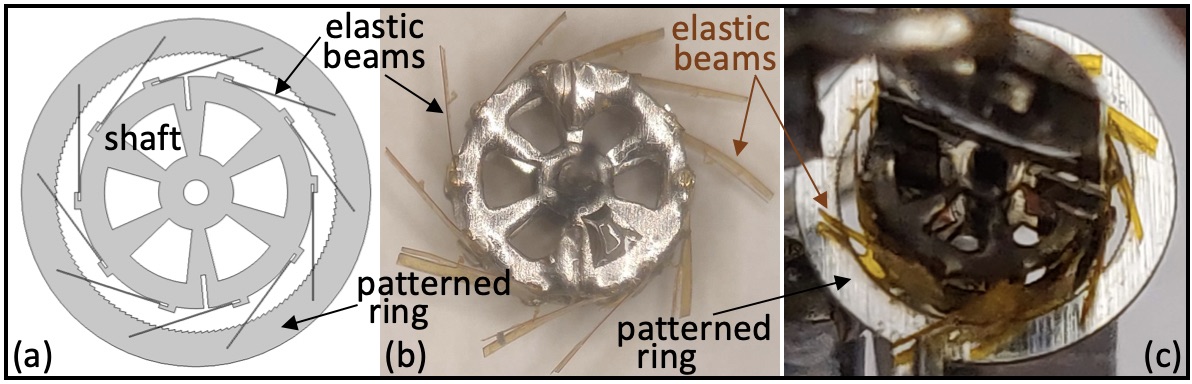,width=3.4in}
\vspace{-1.8em}
\caption{\small{(a) Model ratchet. (b), (c) Fabricated ratchet.}}
\label{fig:ratchet}
\end{figure}

The ratchet is comprised of two parts - an outer patterned ring, and an inner shaft (see Fig. \ref{fig:ratchet}). The inner circumference of the ring is patterned in a zig-zag manner with hills and valleys (that is, is patterned to be rough; see Fig. \ref{fig:ratchet}(a)). 10 elastic beams made from 13$\mu$m-thick Kapton sheet is glued to the inner shaft using 10 laser cut slots on the shaft's circumference. 
When the ring is rotated clockwise, the elastic beams slide freely over the hills and valleys and no torque is transmitted from the ring to the inner shaft. 
When the ring is rotated anti-clockwise, the elastic beams make a head-on contact with the valleys in the pattern effectively locking with the ring and thus torque is transmitted from the ring to the inner shaft. 

The coils are connected to the ring via two rigid rectangular beams as shown in Fig. \ref{fig:parts}(c). The inner shaft is supported using two revolute joints made up of two 0.3mm diameter carbon fiber (CF) rods and two 0.5mm-long Kapton tubes with inner diameter = 0.34mm and outer diameter = 0.4mm. Each CF rod is concentric with one of the Kapton tubes to make the revolute joint. 

It was observed that the ratchet design of \cite{rolling_ral19}, that is, a shaft diameter of 2mm and number of elastic beams = 6, led to high contact pressures at the tip of the elastic beams which wore down the ratchet very quickly leaving the device non-functional. To overcome that, the ratchet was resized to have a shaft diameter of 2.8mm with number of elastic beams = 10 to reduce the contact pressures while transmitting the same amount of torque. 

%%%%%%%%%%%%
\subsection{Steel torsion spring} \label{sec:Steel-spring}

$\theta_{con}$ marks the deflection angle of the steel spring connecting the ratchet's inner shaft to the flywheel (see Figs. \ref{fig:parts}(b) \& (d)).  
Note that due to the anisotropic nature of the ratchet $\theta_{con}$ is always non-negative, that is, it is either zero or positive. 
This implies that depending on the deflection state of the steel spring, it adds either no power or a positive power to the flywheel. 
The details regarding designing and fabricating a steel torsion spring of any given stiffness $k_{con}$ can be found in \cite{baybug18}.
Note that it is crucial that the natural frequency of the steel spring plus ratchet's inner shaft system is $\gg f_{coil} = 250$Hz in order for the steel spring to operate quasi-statically. It is observed that $k_{con} = 75\mu$Nm is sufficient to ensure that, and the resulting spring dimensions are reported in Table \ref{tab:steel-spring}. 

\begin{table}[h]
\normalsize
  \centering 
    \caption{\small Specifications of the Steel torsion spring.}
    \vspace{-0.4em}
    \label{tab:steel-spring}
    \begin{tabular}{|l|r|}
      \hline
      \multicolumn{2}{|c|}{Spring parameters}\\ 
      \hline
      Max. allowable rotation (before failure) & $48^\circ$ \\
      Torsional stiffness $k_{con}$ & 75$\mu$Nm \\
        \hline
      \multicolumn{2}{|c|}{Spring dimensions}\\ 
      \hline 
      Number of parallel beam segments $N$ & 4 \\  
      Beam length $l$ & 2.13mm \\ 
      Beam width $w$ & 0.29mm \\ 
      Beam thickness $t$ & 50.8$\mu$m\\ 
      \hline 
    \end{tabular}
\end{table}

$k_{con}$ was tuned post-fabrication for the correct working of the bot. 
This was done by gluing, and thus effectively eliminating/grounding, some of the beam segments of the steel spring. 
2 out of the 4 beam segments were eliminated this way (see Fig. \ref{fig:parts}(a)) resulting in a spring with twice the stiffness $k_{con} = 150\mu$Nm. 

%%%%%%%%%%%%
\subsection{Full system dynamics} \label{sec:dynamics}

Three sources of torques act on the coil - a restoring torque from the Titanium spring, a torque from the Steel spring, and a torque due to the Lorentz force acting on the coil. The governing equation for the actuator motion can be given by 
\begin{equation} \label{eqn:thcoil}
I_{coil} \ddot\theta_{coil} = \text{--}k_{coil}\theta_{coil} \hspace{0.2em}\text{--}k_{con}\theta_{con} + r\cdot F_{coil}(r\cdot \theta_{coil}, t). 
\end{equation}
During steady-state, the restoring torque from the Titanium spring does not drain or add any mechanical energy to the system and only goes towards maintaining the resonance of the system. The torque from the Steel spring drains some energy away from the actuator every cycle, and the torque due to the Lorentz force adds the same amount of energy back to the actuator every cycle. The way the actuator is constructed, $\theta_{coil,max}$ should not exceed $30^\circ$ else the arm connecting the coil to the Titanium spring will collide with support base of the Titanium spring (see Figs. \ref{fig:parts}(b) \& (d)). Thus $k_{con}$ was tuned post-fabrication to the value of 150$\mu$Nm to ensure that $\theta_{coil,max} < 30^\circ$. 

Three sources of torques act on the flywheel - a torque from the Steel spring, an aerodynamic damping torque due to the motion of the wings, and a torque which models other mechanical losses in the system like the friction in the ratchet and the revolute joints. 
\begin{equation} \label{eqn:thnet}
I_{wing} \ddot\theta_{wing} = k_{con}\theta_{con}  - b\hspace{0.12em}{\dot\theta_{wing}}^2 - \tau_{losses}
\end{equation}
Note that the flywheel always rotates anti-clockwise and thus $\dot \theta_{wing}$ is always positive. During steady-state, the aerodynamic damping torque and $\tau_{losses}$ drain some energy away from the flywheel every cycle, whereas the torque from the Steel spring adds the same amount of energy back. 

$\tau_{losses}$ is set such that it drains $\approx$ 6mW of power, as was estimated in \S\ref{sec:actuator}. The value of the damping factor $b$ is estimated based on the damping torque due to the aerodynamic drag acting on the wing during steady-state. 
\begin{equation} \label{eqn:b}
b\hspace{0.12em}{\dot\theta_{wing, ss}}^2 = \hat p R \cdot F_{D,ss} 
\end{equation}

Lastly, due to the anisotropic nature of the ratchet, $\theta_{con}$ evolves only if either the actuator speeds exceed the wing speed, or when $\theta_{con}$ is already at a non-zero value. 
\begin{equation} \label{eqn:thcon}
        \dot\theta_{con}=
        \begin{cases}
            \dot\theta_{coil} - \dot\theta_{wing} &  \theta_{con}>0\text{, or, }\dot\theta_{coil} > \dot\theta_{wing} \\
            ~~~~~~~0 & \text{otherwise}
        \end{cases}
\end{equation}

\begin{figure}[h] 
\centering
\epsfig{file=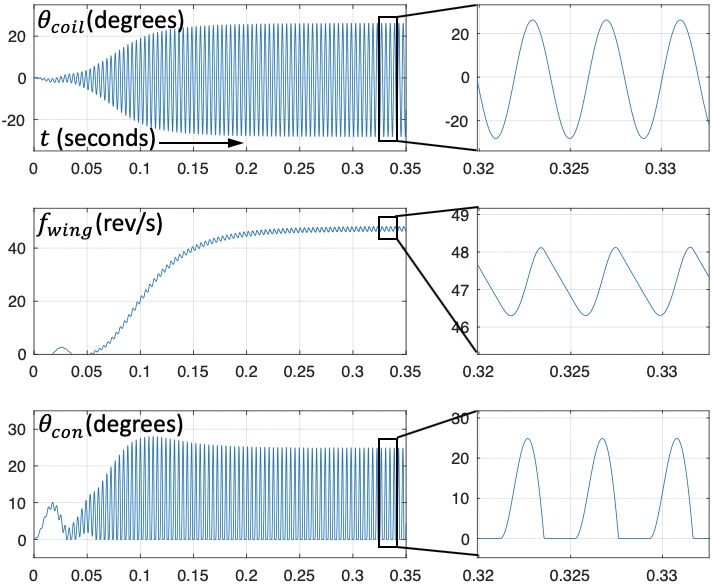,width=3.4in}
\vspace{-1.8em}
\caption{\small{MATLAB simulation of the full system dynamics.}}
\label{fig:sim}
\end{figure}

The full set of system equations were simulated using MATLAB and the results are shown in Fig. \ref{fig:sim}. 
It can be seen that $|\theta_{coil,max}|$ is lower than $|\theta_{coil,min}|$ by $\approx 2^\circ$ since the actuator transmits and thus looses some of its spring potential energy $E_{spring} = \frac{1}{2}k_{coil}\theta_{coil}^2$ every cycle while moving anti-clockwise, that is, when moving from negative $\theta_{coil}$ to positive $\theta_{coil}$. 
Ripples can be seen in $f_{wing}$ after steady-state has been reached showing that the flywheel speed varies by $\approx \pm 2$\% around its mean value of $\approx$ 47.3rev/s as was discussed in \S\ref{sec:flywheel}. Note that $\theta_{con}(t)>0$ when the ratchet is engaged and $\theta_{con}(t)=0$ when disengaged. It can be seen that the ratchet, and thus the actuator, is engaged with the flywheel for $\approx 50$\% of the cycle (that is, for 2ms) and is disengaged in the other half. 

%%%%%%%%%%%%
\subsection{Assembly} 

All the components of the bot were laser cut and then manually assembled with the help of slots that were designed in each of the parts. The weight distribution of the components can be seen in Table \ref{tab:mass}. The support base for the Titanium spring was made as a hollow square box out of 50$\mu$m-thick Aluminum to maximize its stiffness. 

\begin{table}[h]
\normalsize
  \centering 
    \caption{\small Mass distribution of the $\mu$bot.}
    \vspace{-0.4em}
    \label{tab:mass}
    \begin{tabular}{|l|r|}
    \hline
      \textbf{Sub-component} & \textbf{Mass} \\
          \hline
      \hline
      \multicolumn{2}{|c|}{Electrical parts}\\ 
      \hline
      Coil & 13mg \\
      Dead coil & 13mg \\
      Magnet & 24mg \\
        \hline
      \multicolumn{2}{|c|}{Structural parts}\\ 
      \hline
      Titanium torsion spring & 8mg \\ 
      Ratchet's shaft & 7mg \\ 
      Ratchet's ring & 1mg \\ 
      Steel torsion spring & 7mg \\ 
      Wing assembly & 47mg \\ 
      Support base & 13mg \\ 
      \hline
      \hline
      \textbf{Total} & {133mg} \\
      \hline
    \end{tabular}
\vspace{-0.8em} 
\end{table}

%%%%%%%%%%%%%%%%%%%%%%%%%%%%%%%%%%%%%%%%%%%%%%%%%%%%%%%%%%%%%%%%%%%%%%%%%%%%%%%%
\section{Experiments} 

\begin{figure}[h] 
\centering
\epsfig{file=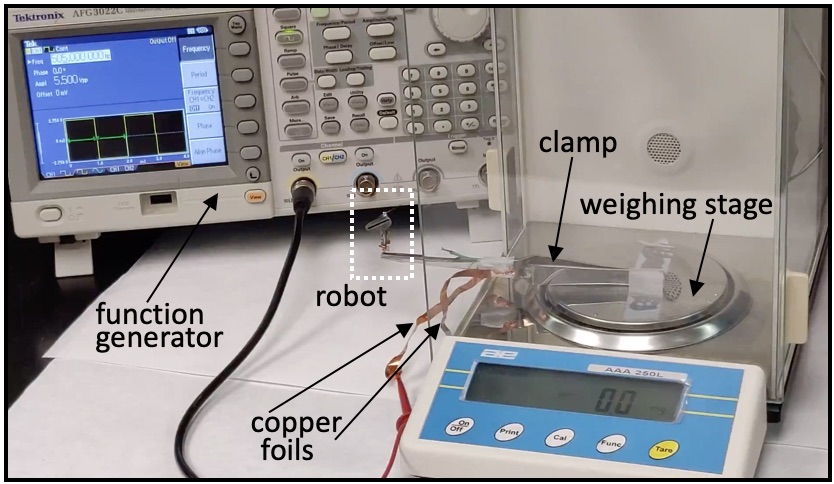,width=3.4in}
\vspace{-1.8em}
\caption{\small{Lift measurement setup.}}
\label{fig:lift}
\end{figure}

The experimental set-up for measuring the lift of the bot is simple and is shown in Fig. \ref{fig:lift}. 
The support base of the bot is clamped through the entire duration of the experiment. 
The clamp is taped to the weighing stage of a 0.1mg resolution weighing scale. 
The 2 thin wires emanating from the coil are electrically connected to 2 $25\mu$m-thin Copper foils which are then connected to the output of a standard function generator. The function generator has some output impedance but it is programmed to only display the voltage applied across a load of 110$\Omega$. 

\begin{figure}[h] 
\centering
\epsfig{file=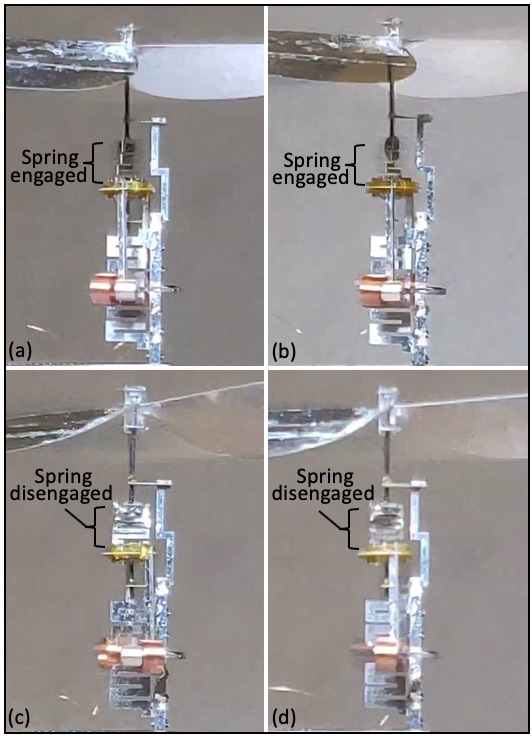,width=3.0in}
\vspace{-0.6em}
\caption{\small{Snapshots of the bot in motion.}} 
\label{fig:snapshot}
\end{figure}

A square-wave voltage is applied using the function generator and resonance is observed at $f_{coil} = 252.5$Hz, which corresponds to the square-wave supply frequency of $2 f_{coil} = 505$Hz. The square-wave amplitude is increased to $\pm 2.75$V, that is, $V_{pp} = 5.5$V to observe a lift of $F_{L,ss}>$ 138mg (= 1.35mN). During this operation, $\theta_{coil, max}$ was observed to be $\approx 26^\circ$ and $f_{wing,ss}$ was observed to be $\approx 47$rev/s. The modeling and simulations done in the previous sections were fitted to these measured values. Fig. \ref{fig:snapshot} shows a few snapshots of the bot while in motion showing the engaged and disengaged states of the spring and the ratchet. 

%%%%%%%%%%%%%%%%%%%%%%%%%%%%%%%%%%%%%%%%%%%%%%%%%%%%%%%%%%%%%%%%%%%%%%%%%%%%%%%%
\section{Conclusions and Future work} 

This work shows that a spinning wing architecture works at the milligram-scale as well, with a lift-to-power ratio comparable to that of the best reported flapping wing bots but at half the net mass. 
If our bot was scaled to weigh 260mg then its lift-to-power ratio will increase by an estimated $\sqrt[3]{2^2}\times$ to 3.6g/W. 
There is still scope to reduce the friction losses in our mechanism to further increase this ratio. 

The novel architecture of our bot offers a few other advantages. 
The actuator motion and wing motion have now been decoupled, that is, there is no need to size the wings depending on the resonant frequency of operation of the actuator. 
The design and cumbersome tuning of the wing hinge for passive wing pitching in flapping wing bots has been eliminated. 
Achieving lower revolutions/second for the wings is now possible while keeping the actuator frequency constant. 
This consumes lower power for generating the same amount of lift since larger wing area and lower rev/s leads to a lower exhaust velocity and a greater efficiency. Lastly, the wings don't need to be light and on the contrary them being heavy is desired thus eliminating the detailed construction of light-weight wings using strong materials like CF veins. 

The wings fabricated for this bot have a high mass of 40mg. Their mass can be reduced by 30mg by shifting the mass towards the wing tips while maintaining the high desired rotational inertia. This saving in mass can be used to add an on-board electronics unit like the one developed in \cite{rolling_ral19} weighing no more than 20mg being low-voltage, and it can be powered using the 8mg photovoltaic cell that was used in \cite{robofly18} to demonstrate untethered operation. 

The use of a spinning wing architecture opens up the possibility of using a quadcopter type of topology for constructing a controllable flying bot in the future. Also, we point the readers to the fact that the design of this bot can in principle work with other small-displacement oscillatory actuators as well. 

%%%%%%%%%%%%%%%%%%%%%%%%%%%%%%%%%%%%%%%%%%%%%%%%%%%%%%%%%%%%%%%%%%%%%%%%%%%%%%%%
\section*{Acknowledgements}

The authors are grateful to get support from Commission on Higher Education (award \#IIID-2016-005)
and DOD ONR Office of Naval Research (award \#N00014-16-1-2206). 
We would also like to thank Prof. Ronald Fearing for his help and insightful discussions.


\begin{thebibliography}{99}

% delfly 
\bibitem{delfly} Delfly. http://www.delfly.nl 

% EM motor scaling 
\bibitem{scaling} W. S. N. Trimmer, ``Microrobots and Micromechanical Systems,'' {Sensors and Actuators}, 19, 267-287, 1989. 

% wood PZT 
\bibitem{best_PZT} N.T. Jafferis, M. Lok, N. Winey, G.-Y. Wei, and R.J. Wood, ``Multilayer laminated piezoelectric bending actuators: design and manufacturing for optimum power density and efficiency,'' {J. Smart Materials and Structures}, vol. 25, no. 5, 2016. 

% wood actuator selection 
\bibitem{actuator_selection} M. Karpelson, G-Y. Wei, and R.J. Wood, ``A Review of Actuation and Power Electronics Options for Flapping-Wing Robotic Insects,'' {IEEE Int. Conf. on Robotics and Automation}, Pasadena, CA, May 2008. 

% robofly 
\bibitem{robofly18} J. James, V. Iyer, Y. Chukewad, S. Gollakota, and S.B. Fuller, ``'Liftoff of a 190 mg Laser-Powered Aerial Vehicle: The Lightest Untethered Robot to Fly,'' {IEEE Int. Conf. on Robotics and Automation}, Brisbane, Australia, May 2018. 

% robobee x-wing 
\bibitem{xwing19} N.T. Jafferis, E.F. Helbling, M. Karpelson, and R.J. Wood, ``Untethered Flight of an Insect-Sized Flapping-Wing Microscale Aerial Vehicle,'' {Nature} 570, 491-495, 2019. 

% zhang EM 
\bibitem{zhang16} Y. Zou, W. Zhang, and Z. Zhang, ``Liftoff of an Electromagnetically Driven Insect-Inspired Flapping-Wing Robot,'' {IEEE Transactions on Robotics}, vol. 32, no. 5, October 2016. 

% lin 
\bibitem{lin_EM1} X. Yan, Z. Liu, M. Qi, L. Lin, ``Low Voltage Electromagnetically Driven Artificial Flapping Wings,'', {MEMS}, pp 1149-1152, Jan 2016. 

\bibitem{lin_EM2} Z. Liu, X. Yan, M. Qi, Y. Yang, X. Zhang, L. Lin, ``Lateral Moving of an Artificial Flapping-Wing Insect Driven by Low Voltage Electromagnetic Actuator,'', {MEMS}, pp 777-780, Jan 2016. 

% wobble motor 
\bibitem{wobble1} S.C. Jacobsen, R.H. Price, J.E. Wood, T.H. Rytting, and M. Rafaelof, ``The wobble motor: an electrostatic, planetary-armature, microactuator,'' {MEMS}, Feb. 1989. 

% RA-L rolling bot 
\bibitem{rolling_ral19} P. Bhushan and C.J. Tomlin, ``An Insect-scale Self-sufficient Rolling Microrobot,'' Robotics and Automation Letters (RA-L) 2019. 

% RA-L jumping bot 
\bibitem{jumping_ral19} [under review] P. Bhushan and C.J. Tomlin, ``An Insect-scale Laser-powered Untethered Jumping Microrobot,'' in review for Robotics and Automation Letters (RA-L) 2019. 

% inchworm 
\bibitem{inchworm02} R. Yeh, S. Hollar, and K.S.J. Pister, ``Single mask, large force, and large displacement electrostatic linear inchworm motors,'' {Journal of Microelectromechanical Systems}, Vol. 11, No. 4, pp 330-336, 2002. 

% flagellum 
\bibitem{flagellum} N. J. Delalez and J. P. Armitage, ``The Bacterial Flagellar Rotary Motor in Action,'' {Methods Mol Biol.}, 1805:33-49, 2018.

% lentink spinning vs flapping 
\bibitem{lentink16} E. W. Hawkes and D. Lentink, ``Fruit fly scale robots can hover longer with flapping wings than with spinning wings,'' {J. R. Soc. Interface}, 13: 20160730, 2016. 

% aerodynamics CL, CD 
\bibitem{passive_rot} J. P. Whitney and R. J. Wood, ``Aeromechanics of passive rotation in flapping flight,'' {J. Fluid Mech.}, vol. 660, pp. 197-220, 2010. 

% iros 
\bibitem{baybug18} P. Bhushan and C.J. Tomlin, ``Milligram-scale Micro Aerial Vehicle Design for Low-voltage Operation,'' {IROS}, Madrid, Spain, Oct. 2018. 

% mems 
\bibitem{baybug19} P. Bhushan and C.J. Tomlin, ``Design of the first sub-milligram flapping wing aerial vehicle,'' {MEMS}, Seoul, South Korea, Jan. 2019. 

\end{thebibliography}
\end{document}